\documentclass[manuscript,nonacm]{acmart}

\usepackage{latexsym}
\usepackage{amsmath}
\usepackage{amsthm}
\usepackage{booktabs}
\usepackage{enumitem}
\usepackage{graphicx}
\usepackage{color}

\AtBeginDocument{%
  }

\title{Transparent Information Flow Modeling for Role-Aware AI Act Assessment}

\author{Mattias Brännström}
\email{mattias.brannstrom@umu.se}
\orcid{0000-0003-3113-2631}
\affiliation{%
  \institution{Umeå University}
  \city{Umeå}
  \country{Sweden}}

\author{Themis Dimitra Xanthopoulou}
\email{themis.xanthopoulou@umu.se}
\orcid{0000-0003-2914-0472}
\affiliation{%
  \institution{Umeå University}
  \city{Umeå}
  \country{Sweden}}

\author{Lili Jiang}
\email{lili.jiang@umu.se}
\orcid{0000-0002-7788-3986}
\affiliation{%
  \institution{Umeå University}
  \city{Umeå}
  \country{Sweden}}

\renewcommand{\shortauthors}{Brännström et al.}

\begin{abstract}
The European Union’s Artificial Intelligence Act introduces a structured framework for governing high-risk AI systems, emphasizing transparency, risk management, and role-specific accountability. However, current assessment practices often rely on monolithic system views or disconnected checklists, limiting their ability to capture the socio-technical complexity of real-world deployments. We explore the use of the Information Flow Methodology (IFM), a modeling-based approach to AI Act compliance assessment. IFM represents decision systems—whether algorithmic, human, or hybrid—as networks of information sites and transformation channels, enabling structured analysis of how information, decisions, and responsibilities flow across organizational and technical boundaries. We apply IFM in a real-world case involving a recruitment support tool used by a public-sector organization,  evaluating both the Provider’s system and the Deployer’s recruitment process. Our results demonstrate that IFM can serve as a scaffold to bridge both component-based and holistic perspectives, and how technical systems are situated within socio-technical contexts. In doing so, IFM supports transparency, identifies risk pathways, and enables alignment analysis between intended and actual use. We argue that such modeling provides a foundational tool for effective, context-sensitive AI governance and accountability—offering capabilities beyond conventional checklist-based assessments.
\end{abstract}



\renewcommand\footnotetextcopyrightpermission[1]{} %
\begin{document}
\maketitle

\section{Introduction} \label{sec:introduction}

The European Union’s Artificial Intelligence Act introduces a comprehensive regulatory framework for governing high-risk AI systems~\cite{EU2024AIAct,EC2024Framework}. It emphasizes transparency, accountability, and human oversight, and defines distinct roles—such as Provider and Deployer—each with formal responsibilities throughout the AI system lifecycle~\cite{EC2024Framework}. While this division of responsibility aligns to some extent with the practical distinction between software vendors and organizational users, it does not always reflect the internal dynamics of AI systems or the socio-technical structures in which they are deployed~\cite{VealeBorgesius2021}.

Assessing compliance under these conditions poses a fundamental challenge: 
how to analyze responsibility, risk, and transparency in ways that reflect both technical architectures and socio-institutional decision-making. Existing technical assessments often treat AI systems as monolithic artifacts — evaluated in terms of model-level accuracy, bias, or fairness metrics—without a structured understanding of how decisions relate to broader socio-technical decision-making process~\cite{green_algorithmic_2020,selbst2019fairness}. This disconnect risks what Selbst et al. have called the \textit{framing trap} —where system-level fairness or compliance is wrongly inferred from component-level properties \cite{selbst2019fairness}. Similarly, the separation of Provider and Deployer provide structurally imposed black boxes which fail to capture the interrelation among use-context, AI system, users, and affected stakeholders~\cite{VealeBorgesius2021}.

In response to these challenges, we have developed a methodology based on information flow modeling~\cite{brannstrom2023impact,ifm2025guide}. This Information Flow Methodology (IFM) models AI-supported decision systems as a set of interconnected information sites (representing states of knowledge or data) and channels (representing decisions or transformations). Together, these components map how information flows from inputs to final decisions. By grounding the analysis in information dependencies and transformations, IFM offers a highly transparent and modular view of both technical components and human processes, including how automation modifies responsibilities and shapes downstream outcomes. The embedding of components within the overall decision process preserves a holistic perspective --- even when focusing on details or tracing the chain of responsibility leading to impacts.

To evaluate the IFM tool in real-world assessment scenarios we applied it in a case of AI Act compliance assessment in collaboration with an external auditor. The case involved a high-risk AI system used to support recruitment decisions in a large public-sector organization. As external assessors, we used the IFM framework as the core of the analytical methodology to examine decision making processes, risks and responsibilities shared and distributed between system developers and the procuring organization. The IFM facilitated the identification of potential risks and biases, and enabled a fine-grained evaluation of whether the system's integration aligned with the requirements of the AI Act.

This paper takes a reflective meta-perspective on this use of IFM in this real-world AI Act assessment. Rather than focusing on the case results themselves, this paper critically examines how the IFM methodological tool performed during the assessment looking into the advantages it offered and the limitation of the approach. %

By addressing these questions, we aim to contribute to the emerging landscape of tools for AI governance. We argue that modeling socio-technical information flows provides a robust foundation for understanding AI system impacts in practice—particularly by revealing how decision elements like models and users interact with formal roles such as Provider and Deployer. This contrasts with monolithic assessments that obscure the realities of how decisions are actually made. 

The remainder of this paper is structured as follows: Section 2 introduces the regulatory background and critiques existing assessment approaches, motivating this study. Section 3 briefly presents the IFM framework. Section 4 describes the real-world use case. Section 5 describes the assessment process and the use of IFM within it to helping to describe integration scenarios and their consequences. Section 6 reflects on methodological contributions, limitations, and directions for future use. %

\section{Background} \label{sec:background}
The European Union’s Artificial Intelligence Act (AI Act) introduces a risk-based regulatory framework designed to govern the AI systems throughout their lifecycle~\cite{EC2024Framework}. The AI Act categorizes systems into four tiers of risk—minimal, limited, high, and unacceptable—with the most stringent requirements applying to high-risk systems. These are systems deployed in sensitive domains such as employment, education, healthcare, and essential public services, where they carry a heightened risk for adverse impacts on fundamental rights. This tight association with fundamental rights occupies a central position in the AI Act and forms the normative anchor for assessment~\cite{EC2024Framework}.

High-risk systems are subject to a range of legal obligations, including requirements for transparency, human oversight, quality management, and risk mitigation~\cite{EC2024Framework}. A core innovation of the AI Act is the distributions of these obligations across the ecosystem of actors involved in its design, development, and use. Most prominently, the Act distinguishes between two primary roles: the \textit{Provider}(typically the developer or vendor) and the \textit{Deployer} (the user organization). This reflects useful real-world asymmetries in access, expertise, control and responsibilities. By separating these roles, the AI Act accommodates both custom development and off-the-shelf procurement, supporting regulatory assessments that reflect the supplier–customer relationships commonly found in practice~\cite{EC2024Framework}.

However, in real-world deployments, the boundary between these roles is rarely clear-cut~\cite{VealeBorgesius2021}. Systems may be adapted or recontextualized after delivery. A single organization may function as both developer and user. Documentation may fail to fully reflect the actual use case. Oversight mechanisms may exist in principle but misalign with the structure of decision-making in practice. Such ambiguities complicate efforts to determine whether obligations are being met—or even whether they are assigned to the correct party~\cite{green_algorithmic_2020}. Effective assessment must be capable of navigating this hybrid, evolving, and often ambiguous landscape.

Yet rather than engaging with this complexity, many current assessment frameworks treat AI systems as monolithic entities~\cite{selbst2019fairness}. Technical evaluation methods tend to focus narrowly on the model in isolation, using performance metrics or fairness measures derived from datasets and outputs~\cite{selbst2019fairness}. Such metrics are often applied without consideration of the socio-technical environment in which the system is embedded—even though it is precisely within this context that stakeholder impacts ~\cite{binns2018fairness}actually materialize. 

A similar abstraction occurs in policy and governance frameworks, where AI ethics guidelines and regulatory templates frequently adopt a developer-centric perspective or assume a linear pipeline from design to deployment. This perspective obscures the distributed and context-dependent nature of responsibility in real-world deployments. As a result, many frameworks struggle to separate roles, trace obligations, or adapt assessments to hybrid configurations. Examples of such ``monolithic'' approaches include tools like the European Commission’s ALTAI (\textit{altai.insight-centre.org}), the Dutch FRAIA template\footnote{https://www.government.nl/documents/reports/2021/07/31/impact-assessment-fundamental-rights-and-algorithms}, and the Swedish DIGG trust model (\textit{digg.se}).

These limitations have been widely critiqued in the literature. 
Veale and Borgesius argue that role separation in regulatory frameworks, such as the distinction between Provider and Deployer in the AI Act, fails to cleanly reflect the realities of distributed responsibility and dynamic system development~\cite{VealeBorgesius2021}. Metcalf et al. emphasize that accountability must be tied to the actual sites of impact within socio-technical systems, not be abstracted through technical compliance~\cite{metcalf2021algorithmic}. Weinberg et al. caution that fairness metrics cannot substitute for attention to concrete social consequences~\cite{weinberg2022rethinking}. Green and Viljoen critique the foundational assumptions of algorithmic formalism—namely, that algorithms can be neutral, evaluated in isolation, and applied without contextual sensitivity (universalism)~\cite{green_algorithmic_2020}. In algorithmic fairness, in particular, passing a fairness test at the model level is often mistakenly equated with overall fairness in a socio-technical system \cite{selbst2019fairness, weinberg2022rethinking}. This fallacy --- termed the \emph{framing trap} by Selbst et al --- mistakes properties validated in an algorithmic frame to hold in the socio-technical frame. More broadly, treating compliance as an abstract property of a technical artifact undermines accountability and risks disconnecting obligations from the decision-making processes in which harms emerge.
Transparency is sometimes raised as a general solution, without sufficient specification on what is made transparent, to whom and how. Ananny and Crawford argue that genuine accountability requires understanding how information is used within specific context—not merely exposing inner workings~\cite{ananny2018seeing}.

Taken together, the \textit{framing trap} at the technical level and the monolithic assumptions at the organizational level pose significant challenges for responsible AI governance. Achieving meaningful compliance under the AI Act requires more than token documentation or fairness metrics. It demands a grounded understanding of how decisions are made across the system as deployed: how information flows, who takes actions, what transformations occur, and how these elements connect to obligations and real-world impacts. 
A number of frameworks and initiatives have begun to address this challenge, though most still face limitations in scope or applicability.

\subsection{Related Work}

In response to these challenges, several frameworks aim to bridge the accountability gap in AI systems by tracing roles and responsibilities across the lifecycle. Raji et al.\cite{raji2020closing} and Cobbe et al.\cite{cobbe2021reviewable} propose internal auditing and reviewability models grounded in documentation and role separation, while Lopez~\cite{lopez2021bias} and Cobbe et al.\cite{cobbe2023supply} highlight structural origins of harm through bias typologies and distributed accountability. These approaches foreground institutional dynamics and socio-technical complexity but treat technical systems largely as opaque components. Novelli et al. propose scenario-based and proportional risk assessment methodologies under the AI Act\cite{novelli2023taking,novelli2024scenario}. While these incorporate contextual sensitivity, they do not model socio-technical decision structures or trace how automation reshapes accountability across system layers. A more technical perspective is offered by OMNI~\cite{dignum2005omni}, which models agent roles and normative constraints within multi-agent systems. While structurally expressive, OMNI remains focused on prescriptive agent design and does not readily apply to socio-technical assessment. Together, these contributions underscore the need for structurally grounded methods that can model how automation reshapes decision-making, responsibility, and impact across both technical and social domains.

\subsection{The Need for Structural and Context-Aware Methodologies}

These limitations reveal the need for methodologies that capture both the technical properties of AI models and the structure of decision-making across socio-technical systems.

Together, these critiques highlight the need for a methodological scaffold that supports both granular and holistic assessment.
In the following we list key desirable properties tied to critique and needs.

\begin{enumerate}
    \item \textbf{Situate decisions within institutional and social context} \\ \emph{To:} Capture how decisions are made within the broader organizational and normative context in which a system operates—reflecting how roles, constraints, and institutional practices shape outcomes~\cite{green_algorithmic_2020, wieringa2020account, selbst2019fairness, metcalf2021algorithmic, ananny2018seeing, binns2018fairness}.

    \item \textbf{Describe roles, responsibilities, and how they shape decision-making} \\ \emph{To:} Make visible who is involved in shaping, deploying, or relying on a system, what responsibilities they hold. Show how these responsibilities are distributed over the socio-technical decision-context~\cite{VealeBorgesius2021, metcalf2021algorithmic, selbst2019fairness, binns2018fairness}.

    \item \textbf{Interpret risks and impacts in relation to context-specific norms and social consequences} \\ \emph{To:} Link behaviors to their situated effects, recognizing that harm and acceptability depend on domain norms, power dynamics, and institutional context~\cite{metcalf2021algorithmic, green_algorithmic_2020, weinberg2022rethinking, selbst2019fairness, binns2018fairness}.

    \item \textbf{Support structural comparison, nesting and compositional modeling} \\ \emph{To:} Accommodate modular and evolving systems, enabling comparison across deployments, and tracing how decisions and responsibilities emerge across interconnected components and their use contexts~\cite{VealeBorgesius2021, green_algorithmic_2020, metcalf2021algorithmic}.
\end{enumerate}

A structural perspective is essential to move beyond tokenistic assessments. As Metcalf et al. and Selbst et al. argue, impact and risk should not be treated as isolated outcomes but understood as structurally linked to roles, obligations, and decision processes---visible and analyzable across the system~\cite{selbst2019fairness,metcalf2021algorithmic}.
This perspective is particularly urgent in the context of the EU AI Act, which grounds its risk-based regulatory framework in the protection of fundamental rights. 

This paper examines the use of a methodology to address these concerns: \textit{Information Flow Methodology} (IFM), a modeling framework developed to transparently trace bias and impact in socio-technical decision-making~\cite{ifm2025guide}. 

In this study, we explore how using the IFM can support a structured AI Act assessment of Provider and Deployer obligations. Here IFM serves as a foundation for describing socio-technical decision-structures across algorithms, decision-makers and organizational contexts. This approach  help clarify the relationship between Providers and Deployers, reveal mismatches between intended and actual use, and make visible the structures that support or undermine accountability and compliance without focusing solely on isolated components, nor obscuring the essential decision-structure.

\section{The IFM Framework} \label{sec:framework}

    Studying how decision-making occurs in socio-technical systems requires a clear and systematic representation of how information flows through the decision process. The Information Flow Methodology (IFM) offers a structured way to describe these processes, enabling the identification of potential biases and the evaluations of how outcomes affect stakeholders.
    The IFM model was developed inspired by the works of Barwise \cite{barwise1995logic, barwise1997information} and uses similar components.
    
    This section first provides a conceptual overview of the IFM and its core components (Section \ref{sec:IFMdescription}), followed by descriptions on the two structured properties - bias and impact, in IFM (Section \ref{sec:biasInIFM}).

    A more comprehensive description of the IFM framework can be found in the IFM Method guide~\cite{ifm2025guide}.
    
    \subsection{An Overview of the IFM Model}\label{sec:IFMdescription}

    The IFM Model provides a detailed description of the decision-making process by specifying the information available for each decision and the resulting outcome. This specification enables a thorough analysis of the factors that might influence a decision outcome and how the resulting outcomes might, in turn, impact other decisions.

    The IFM consists of three primary components:
    \begin{enumerate}
        \item \textbf{Sites:} Represent the states or repositories of information within a decision process, such as inputs, intermediate results, or outputs.
        \item \textbf{Channels:} Represent the transformations, operations, or decisions that move information from one site to another.
        \item \textbf{Networks:} Represent sites and channels connected to form directed graphs, capturing entire decision processes from initial inputs to outcomes.
    \end{enumerate}
    
    Any decision process that begins with some initial information and ends with a decision can be described using these elements. The decision process itself constitutes a network, each sub-decision a \textbf{channel}, and the inputs, output decision, and any intermediate result would be \textbf{sites}.

    In diagrams, sites are typically text labels, and channels as arrows connecting them. The direction of the arrow follows the flow of information—from upstream inputs to downstream results. Channels can be presented one by one with their sites or connected together into a network graph. Sites only participating in a flow as reference material or other assisting information to a channel can be marked as \emph{[ site ]}, such sites are termed as \emph{static} sites~\cite{ifm2025guide}.
    
    \newpage 
    
    \emph{Example:}
    
    \[
    \begin{array}{c}
    \text{[Eligibility Criteria]} \\
    \text{Application}
    \end{array}
    \xrightarrow{\textit{Eligibility Check}}
    \text{Eligibility Result}
    \]
    
    Here, \textit{Eligibility Check} represents a channel that takes the \emph{Application} as input and produces the \emph{Eligibility Result} using the static \emph{Eligibility Criteria}~\cite{ifm2025guide}.

    Following this conceptual overview, we extend the model with a formal definition to enable more rigorous analysis.

    \subsubsection{Formal Definition of the IFM Model} \label{sec:ifm_formal}

    A Network $N = <S, C, T_S, T_C, \mathcal{T}, \mathcal{R}>$ where $S$ is a set of Sites representing sources of information, $C$ is a set of Channels representing directed transformations between sites. $T_S: S \rightarrow T$ and $T_C: C \rightarrow T$ are functions assigning \emph{types} to sites and channels respectively. $\mathcal{T} = (T, \leq_T)$ is a type system where $T$ is a set of type classes and $\leq_T$ is a partial ordering defining subtype relationships. $\mathcal{R}$ is a set of inference rules determining properties and relationships based on types.

    A Site $S_i \in S$ represents a source of information on which a decision can be made. While sites are unique in $N$, the information they contain can be further described by $T_S$ assigning them one or more types. Types will be further described below and are not unique to a site.
    A Channel $C_j \in C$ is a directed transformation between one or more input sites $In(C_j) \subseteq S$ with one or more output sites $Out(C_j) \subseteq S$. Like sites, channels are unique in $N$ but can be assigned channel types via $T_C$ or inferred types via $R$.

    A valid network $N$ forms a directed acyclic graph (DAG) where:
    \begin{itemize}
    \item A site in $S$ can be the output of at most one channel in $C$.
    \item The output of a channel cannot serve as an input to an upstream channel (no cycles).
    \item $In(N) = \{S_i \in S \mid S_i \text{ is not an output of any } C_j \in C\}$.
    \item $Out(N) = \{S_o \in S \mid S_o \text{ is not an input to any } C_j \in C\}$.
    \item $Mid(N) = S \setminus (In(N) \cup Out(N))$ \quad (intermediate outputs of $N$).
    \end{itemize}
    
    A network $N$ can be seen as an abstract channel $N: In(N) \rightarrow O$ where $O \subseteq Out(N)$. Similarly, any channel $C$ can be further defined as a network. This allows a network $N$ to be nested within another larger network $N'$ where it participates as a channel. This nesting provides the possibility of abstraction, as parts of large or complicated networks can be abstracted as channels representing sub-networks. It also allows causal reasoning about the properties that networks have as channels. A refinement order over such networks, and a formal account of which stakeholder questions a network can answer, is developed in~\cite{brannstrom2026answerability}; here we use the networks informally as a modeling language.

    \subsection{Bias and Impact}\label{sec:biasInIFM}

    In this framework, \emph{bias} refers not to normative unfairness, but to a structural deviation from the intended or declared function of the decision process. This operational framing enables analysis of how such deviations propagate through socio-technical systems. \emph{Impact}, by contrast, refers to the potential real-world harms such deviations may cause—particularly in relation to rights and fairness.
    
    In the IFM model, both \emph{bias} and \emph{impact} are treated as \emph{structural properties} of the modeled decision process. They reflect different but connected forms of misalignment that may affect stakeholders or violate regulatory norms and obligations. In line with~\cite{metcalf2021algorithmic}'s critique on impact assessments, \emph{impacts} in this framework do not refer merely to statistical deviations or generalized risks, but to structurally accountable relationships between actors and potential or realized harms.
    
    Separating \emph{bias} and \emph{impact} clearly, conceptually and structurally --- one reflects decision properties, the other downstream normative effects --- is essential to avoid the framing trap. Treating fairness as a property of isolated model outputs, risks obscuring the socio-technical pathways through which biased decisions become real-world harms~\cite{weinberg2022rethinking, selbst2019fairness,binns2018fairness}. Here we instead structurally separate concepts and reveal these pathways of harm.

    \textbf{Definition of Bias:} Bias, in this context, refers to a systematic deviation from the described function of the decision process. Such biases might result in decreased accuracy through false positives and false negatives and the effects might be limited or disproportionate to specific subsets of input, such as favoring particular groups.
    Bias may propagate \emph{downstream} through dependent channels and sites. If its influence is halted or unused, the propagation path ends.
    This makes it possible to trace the scope and effect of bias and mitigating efforts across the network, in a transparent and formally bounded way.
    
    \textbf{Definition of Impact} here refers to the real-world normative consequences or risk thereof of decisions, especially unintended harms affecting stakeholders or sub-groups. While all decision outcomes can be said to produce impacts, we here focus on \emph{unintended harmful impacts}. These impacts strongly relate to bias as bias are sources of unintended decisions in the information flow. 
    Impacts appear at downstream ends of the DAG as stakeholder outcomes. Understanding their causes requires tracing paths \emph{upstream} towards contributing decisions.
    Where downstream bias and upstream impact paths intersect, bias may a contributing cause to the impact.
    
    This enables a structured comparison between \emph{declared responsibilities}—such as those stated in compliance documentation—and the actor’s actual \emph{operational influence} over decision-relevant information or processes, helping to identify gaps between legal obligation and effective control.

    The ability to explicitly model these paths enables the identification of critical control points and the evaluation of whether obligations such as human oversight, transparency, or robustness have been fulfilled.

\section{The Use Case}
\label{sec:use-case}
To evaluate the practical value of IFM, we collaborated with an independent auditor that conducted a pre-deployment assessment of an AI-supported recruitment support tool for a public-sector municipality. The municipality, preparing for the upcoming EU AI Act, sought to understand how such systems could be assessed under the regulation, which classifies recruitment tools as high-risk due to their potential impacts on employment and fundamental rights. The work is part of a larger ongoing effort by the municipality around responsible AI in public-sector procurement and development.

The system—developed and maintained by a private AI vendor—had already been adopted in several organizational settings internationally. It provides interview-based screening, candidate scoring, and shortlisting functionality through an AI-powered service platform. The assessed deployment scenario involved integrating this system into the municipality’s existing recruitment process. Although the Provider had articulated a commitment to legal and ethical compliance, the system had not yet undergone external review with respect to AI Act requirements.

The assessment aimed to examine not only the system’s documentation and declared properties but also how its integration might reshape responsibilities, risks, and decision-making structures in the local institutional context. This setting offered a valuable real-world test for the IFM approach, combining high socio-technical complexity with distributed decision-making and multiple possible integration paths. Both the Provider and the Deployer were committed to responsible AI use and engaged openly, providing access to documentation and stakeholder interviews.

Several distinct stakeholder groups are involved. %
For clarity, we summarize these roles below:

\begin{itemize}

    \item \textbf{Provider:} The organizational entity that develops and places the AI system on the market, as defined by the AI Act. In this case, the Provider offers a recruitment support tool as a software service.
    
    \item \textbf{Deployer:} The organization that integrates and uses the AI system in its operations. Here, the Deployer is the municipality, a public-sector organization with large recurring recruitment needs.

    \item \textbf{Recruiter:} Internal staff (e.g., HR specialists) who interact directly with the AI system to evaluate candidates, interpret outputs, and participate in hiring decisions.

    \item \textbf{Candidate:} Individuals applying for positions, who are evaluated through processes that may involve both automated and human-led assessments.

    \item \textbf{Client / Recruiting Manager:} The internal organizational unit or manager initiating the recruitment—i.e., the one with the vacancy to fill. The client is the ultimate recipient of the recruitment outcome. In this case, the client is the \textbf{Recruiting Manager} of the relevant unit, who holds formal responsibility for assigning the position to a candidate.

\end{itemize}

The next section describes how the IFM was used in practice to model the Provider’s system, the Deployer’s recruitment workflow, and the integration scenarios under consideration.

\section{A Structured, Role-Sensitive Assessment Methodology}
\label{sec:assessment-methodology}

The assessment involved three interlinked parts:

\begin{enumerate}
    \item analysis of the Provider and the AI tool itself;
    \item analysis of the Deployer’s pre-integration recruitment process;
    \item modeling and evaluation of potential integration scenarios.
\end{enumerate}

Each stage used participatory IFM modeling to describe decision processes, identify risks, and support AI Act compliance evaluation.

Below, we describe each of these elements and the stepwise procedure used to elicit IFM models and apply them in support of AI Act compliance assessment.

Due to confidentiality, we do not cite stakeholder documents or interviews directly. Insights are presented through the modeling process, with focus on method over case outcomes.
Some details are omitted for anonymization. 

\subsection{IFM Step-Wise Methodology} \label{sec:ifm-method}

The IFM modeling process follows a stepwise process to construct and analyze socio-technical decision structures. These steps serve both technical modeling and participatory engagement.

For a comprehensive treatment of the methodological steps, see the IFM Guide~\cite{ifm2025guide}. The steps are briefly summarized below:

\begin{enumerate}
    \item \textbf{Sketching.} Here the decision process is captured as informal sketches. The sketches identify key process stages, inputs, and outputs in direct collaboration with stakeholders, from interviews, or from documentation.
    
    \item \textbf{Refinement.} Sketches are consolidated and refined into a formal IFM network, specifying decision points (channels), information states (sites), and their interdependencies.
    
    \item \textbf{Types and Roles.} Here each site and channel is assigned semantic types and descriptions clarifying what transformation is happening at each step. Each site, channel or sub-network is associated with a responsible actor or role in the process.
    
    \item \textbf{Bias and Decisions.} By examining each transformation, potential local biases and problems can be detected. Key properties of decision-making are identified, including transformation types and potential bias propagation. %
    
    \item \textbf{Stakeholders and Risk.} Here decision points and outcomes are connected to possible affected stakeholder-groups and potential downstream impacts. This forms the basis for analyzing interaction between decisions, bias and impact.
    
    \item \textbf{Analysis} The model is used to ground evaluation of key properties such as transparency, oversight, traceability, and alignment with declared responsibilities or regulatory obligations.
\end{enumerate}

In the use case, the six-step procedure was applied independently to both the Provider’s AI tool and the Deployer’s recruitment process in a slightly adapted form based on the available information and stakeholder access. A final integration phase then combined these models to evaluate their alignment and to support compliance assessment under the AI Act. The following subsections describe each phase in turn, highlighting how the method was realized in practice.

\subsection{Analysis of the Provider}
\label{sec:provider}

The first step focused on the Provider’s recruitment tool, with two aims: evaluating its compliance efforts and analyzing how its risks affect the Deployer’s decision-making and risk landscape.

To address these goals, the assessment started by examining the decision process embedded within the Provider’s system, its intended use, and its classification under the AI Act. As the system operates in a high-risk domain—recruitment—the Provider had classified the tool accordingly, identifying it as a high-risk AI system. However, the Provider also emphasized that the \emph{intended use} of the tool is solely as a support for recruiters, not to by itself make recruitment decisions.

To gain an understanding of how the tool functions, the first step was to construct an IFM model using the methodology described in section \ref{sec:ifm-method}. The \textbf{sketching} step (1) of the method were carried out using a combination of sources: extensive documentation provided by the Provider, a live system demonstration, and four interviews—two with representatives of the Provider and two with employees of the Deployer who had previously encountered the tool as candidates in a different organizational context. Together, these sources provided a multi-perspective view of the system’s components and its abstract decision logic. The material from this sources were captured as part of the note taking involved where the sketched IFM structure served naturally to categorize the information. The sketched models from the notes were consolidated and refined as per steps 2 and 3 resulting in the model described below.

\subsubsection{Information Flow Structure}

\begin{figure*}[htbp]
    \centering
    \includegraphics[width=\textwidth]{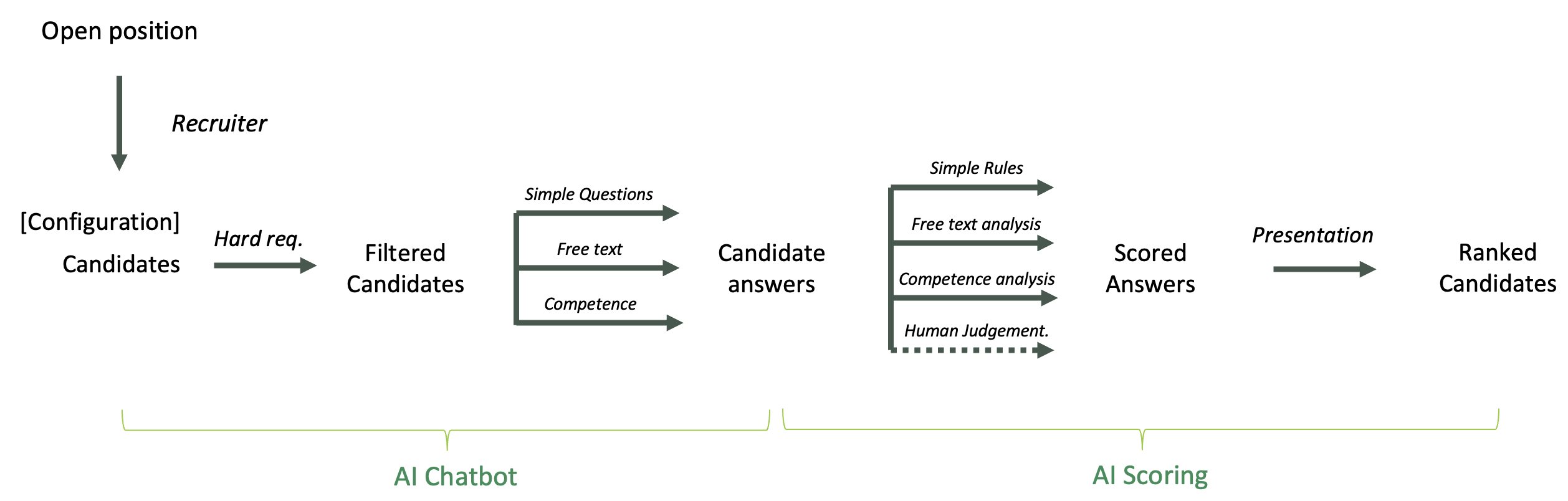} %
    \caption{An abbreviated version of the information flow model of the Provider's AI system for recruitment. Inputs are a Client recruitment request, captured as a requirement configuration, and a set of candidates. The output is a list of ranked and scored candidates.}
    \label{fig:ifm tool}
\end{figure*}

The system was found to consist of two core components, each aligned with the configuration of requirements tailored to the job position. The first is a chat-based interview module that interacts with candidates by presenting a series of structured questions in a conversational format. This module does not evaluate the responses but simply records them.

The second component is the scoring module, which analyzes the responses to each question and assigns numerical scores. Scoring methods vary by question type—some use rules, others rely on semantic models tailored to each category.
In certain cases, scoring is instead delegated to outsourced human assessors. The final score for each candidate is calculated by a weighted summing the individual question scores, which is then used to generate a ranked list of candidates.

The IFM model captured this process as a structured network of information sites--such as candidate inputs, model scores---and transformation channels, including the candidate answering, the evaluation models, etc. This model clarifies how recruiter-facing outputs are generated. An abbreviated version of the IFM model is shown in Figure \ref{fig:ifm tool}.

\subsubsection{Bias and Risk}

Following the construction of the IFM model, the model itself was analyzed for bias and problems following step 4 and 5 of the IFM method. Looking to the abstract information flow by itself does not confirm the existence of biases but shows where there is a potential for them unless mitigating factors are at play. This step was done before evaluating the Provider's own risk management material in order to later see how well it lined up. Below, a few key observations of this step are highlighted:

\paragraph{Evaluation biases}

Potential bias is assessed per channel by analyzing inputs, outputs, and transformation type.

In the evaluation of free-text material, especially in a high-risk area such as recruitment, the potential for group-specific biases is of high concern~\cite{raghavan2020mitigating}. The granular approach, where each question is scored independently limits the potential for such biases. Consider the information flow for the scoring of a single question:
$$
\textit{Response} \xrightarrow{\textit{translation}} \textit{Normalized} \xrightarrow{\textit{Scoring model}} \textit{Sub-score}
$$
At each scoring instance, no contextual information besides the direct answer is used. All answers go through translation, diminishing the potential for language-based discrimination including on grammar and spelling due to the partial normalization and loss of information. The potential bias is further reduced as each model is fine tuned to a particular type of question. Separate scoring isolates potential bias to each question, unlike holistic interview scoring. 

Therefore, based on the information flow model alone, the potential for group-specific discrimination in scoring appeared low. This was also later supported by the Provider's own statistical bias assessment tests showing no detectable bias on gender or ethnicity in the most used set of scoring models. This however only holds on the scoring-component level and does not entirely rule out system level discriminatory effects.
Of particular note here is that the fallback human evaluation does not operate on separate questions but on the whole interview. Since human scoring operates on the full interview rather than individual questions, the mitigating factors described above (decontextualization, normalization, and bounded scope) do not apply to the same degree. This increases the potential for subjectivity and bias towards protected groups as the human assessors can access more contextual cues and cross-interpret answers.

\paragraph{User-facing Transparency}

Transparency toward Deployers is mandated under Article 13 of the AI Act, including the requirement to provide sufficient information about the system’s functioning and outputs to enable meaningful oversight and risk assessment~\cite{EC2024Framework}. Transparency and explainability can be assessed in two related ways using the IFM model. To begin with, the information present on the sites, semantically contextualized by the channels between them represent information present in the system (down to the abstraction level of the model). We can ask on one side, for transparency: Is this information made available to an observer of the process? And on the other, for explainability: Is this information enough to cover the information needs of the observer? Both questions are strictly contextually dependent on who the observer is. 

The granular treatment of the interview --- where each question is scored separately --- means that detailed information leading up to the final score \emph{is} available in the system. This design enables high transparency. As most of this information—such as each question, chat answer, and corresponding score—was made available to the recruiter in the interface, much of this transparency was also realized.   
However, the variable source of the scoring --- AI model or human — was \emph{not} disclosed. This lack of visibility introduces an opacity that may be relevant for oversight and accountability, particularly where users expect consistent evaluation. This is of particular concern in a high-risk setting, as human evaluation operate on different information and carry different potential for bias. Key to note here is that the IFM highlights that this information \emph{is} present in the system, but \emph{not} disclosed to the user.  

\paragraph{Candidate-facing Transparency}

It becomes clear from the IFM model that the interview process and the scoring are performed by entirely separate models. This, however, was not clear to the candidates we interviewed. While they were informed that they were interacting with, and were to be evaluated by, AI, they were not informed of this separation. This is relevant since the chat experience is intentionally designed as a natural conversation. While answers theoretically could have been collected by other means, like a form, the natural language chat improved candidate retention and captured attention. This exemplifies the Eliza effect: candidates attribute social norms to the human-like interface~\cite{nass2000machines,weizenbaum1966eliza}.
During the interaction, the candidates are given the impression that this perceived conversational entity is also the one evaluating their interview, when in fact the evaluation happens using an entirely disconnected mechanism using far less conversational models.
An additional, opposite variant of the same problem is the potential human scoring --- the candidates are told they will be interviewed and evaluated using AI, not that they will be by a human. In these cases, if the IFM we are discussing here, or the equivalent information, were disclosed to the candidates, the above potential issues would be reduced --- marking them as transparency problems/biases (bias to the candidates behavior due to lack of information).

\paragraph{Automation, Agency and "Human-in-the-Loop"}

Automation bias is the tendency for human decision-makers to over-rely on automated systems—accepting their suggestions even when incorrect—due to misplaced trust, reduced vigilance, or interface design that structurally encourages deferral~\cite{parasuraman1997humans, dzindolet2003role,goddard2012automation}.
Automation bias and effective agency can be assessed in the context of IFM using the following general model of decision-making.
$$
    I \xrightarrow{G, A} O
$$
where $I$ represents the information an agent, the user, has available to make their decision. $O$ represents the affordances, choices, the agent has available. $G$ represents the governance or contextual constraints as well as the decision mandate the agent operates under. What remains for $A$ is the effective agency envelope under which the agent can make responsible decisions. Factors such as incomplete information ($I$), uneven cognitive workload across options ($O$), limited decision mandate, or rigid procedural constraints ($G$) can all restrict the agent’s remaining effective agency ($A$). A full exposition of this approach to agency evaluation is out of scope for this paper and will be detailed elsewhere.

In the declaration of the system's \emph{intended use}, the Provider is clear that the system is scoring the candidates, but not making decisions on who goes on to be hired or not. This is up to the recruiter. The final output of the system is presented in the recruiter's Applicant Tracking System (ATS) as a ranked and scored list. Each candidate can then be further opened and inspected down to individual questions, as previously mentioned. There is, however, no functionality to edit the scores, and thereby change or actively work with the ranking. This raises the potential of automation bias --- in a high-volume recruitment scenario, the recruiter might not have the time to manually go through all the answers. And even if they do --- if they do not have tool support to modify the scoring and thereby co-create the final ranking --- it becomes much more work intensive to take a sufficient degree of responsibility compared to just accepting the AI answers. An imbalance of this sort on user affordances drives automation bias~\cite{goddard2012automation}. Unhandled, the presence of this concern makes automation bias a risk and also makes the \emph{intended use} itself potentially problematic and in need of address for the Deployer.

\subsubsection{Grounding Risk and Compliance Analysis}

With the IFM model and analysis as a backdrop, an analysis of the Provider’s declared risk mitigation strategy and compliance claims was then conducted. During this process, potential bias found in the model can be addressed if their path to having impacts is blocked by risk-mitigation efforts or tests which demonstrate the absence of bias. 
The Provider’s risk assessment aligned with many model findings but tended to overemphasize model-level properties,
such as their accuracy and fairness measurements, rather than how they contributed together with socio-technical components to the final decisions. 
Hybrid components ---like human scoring fallback and automation bias --- were not fully addressed by tests or risk management. The IFM model thus served as a tool to identify coverage gaps in the risk analysis, including responsibilities not clearly assigned or residual risks not accounted for.

Finally, the model helped clarify how responsibility for remaining risks --- particularly automation bias and candidate misunderstanding -— might shift toward the Deployer in practice. This insight informed the subsequent modeling of the Deployer’s recruitment process and the evaluation of how integration scenarios might amplify or mitigate these residual issues.

This analysis of the Provider is incomplete without the Deployer perspective, since actual impacts only materialize where the system is deployed. Only by tracing the full chain from decisions to affectees can risks and impacts be assessed meaningfully~\cite{selbst2019fairness,metcalf2021algorithmic}. Sections \ref{sec:deployer} and \ref{sec:integration}, examine the Deployer’s process and effects of integration choices.

\begin{figure*}[htbp]
    \centering
    \includegraphics[width=\textwidth]{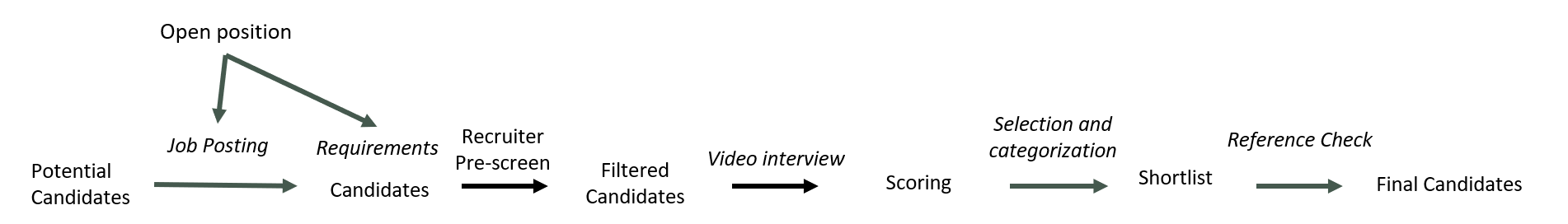} %
    \caption{An abbreviated version of the information flow model of the Deployer's recruitment process, prior to AI tool integration.}
    \label{fig:ifm volume}
\end{figure*}

\subsection{Analysis of the Deployer pre-integration} \label{sec:deployer}

The second step analyzed the Deployer’s recruitment process prior to tool integration, focusing on how decisions were made and on what basis.
A stakeholder analysis was also conducted to identify affected parties, decision-makers, and responsibility holders.
This included how intermediate decisions where logged and communicated internally as this determines how chains of accountability can be established and traced.

The Deployer’s recruitment practices were modeled through ten stakeholder interviews—spanning recruiters, HR personnel, technical staff, and recruiting managers—and a review of internal guidelines and documented procedures. The organization maintained extensive documentation addressing fair and equal treatment of candidates, though operational practices varied depending on recruitment context. Stakeholder perspectives on AI differed, but most emphasized the importance of maintaining human control and fairness in recruitment decisions. A strong sense of personal responsibility for ensuring a fair process was consistently expressed, aligning with the organization’s documented guidelines.

In contrast to the Provider analysis, interviews were the primary source of information for constructing IFM models of the Deployer environment. The IFM modeling process fit naturally within the interview context. Sketches were compiled during note-taking, enabling a form of partial participatory co-design, with the assessor responsible for integrating and refining the models.

Unlike the Provider’s technical focus, Deployer models were predominantly socio-technical, representing the recruiters as individuals or groups operating at decision points in their professional capacity, partly coordinated through their Applicant Tracking System. A key feature of the IFM approach is its ability to represent both human and automated decision systems using a unified modeling language.

IFM makes visible what information is available at each decision point --- key for understanding how decisions are formed and justified. In the recruitment process, the semantic character of the decision-system can shift significantly depending on the order in which decisions are made: Sequence changes affect available evidence and thus alter decision logic. This is of particular importance when considering the effects of integrating AI tools or replacing components with automation, as each such modification can lead to the end decision being taken on completely different grounds.

While recruitment generally followed a standardized procedural framework, adaptations were made for different recruitment contexts. Two IFM models were constructed to reflect this variation: one for \emph{executive recruitment} and one for \emph{high-volume recruitment}. 
The executive process involved a multi-step, collaborative evaluation, while the high-volume track use video-based interviews and a sorting process optimized for throughput.
While both of these tracks were modeled and analyzed in depth in the assessment process, and both present challenges worth exploring --- this paper focuses on the high-volume track.

\paragraph{The high-volume recruitment} This version of the process was used for operational roles with large applicant pools. This process was designed for throughput, emphasizing efficiency and role-matching. An abbreviated IFM model of this process can be seen in Figure~\ref{fig:ifm volume}.
Central to this process is a self-recorded video interview which is subsequently assessed by recruiters. Interviews revealed that the video assessment plays a structurally central role for initial candidate impressions. While not necessarily problematic, this placed significant evaluative weight on visual and linguistic presentation, potentially exposing the process to recruiter bias.
In contrast to the Executive track the high volumes necessitated more formal scoring and record-keeping of candidate attributes as  main carrier of information between decisions. These records, and searching over them were especially important for the sorting phase where candidates were allocated by recruiters based on attributes and preferences to suitable open positions.
The reliance on recorded scores and attributes makes oversight harder, but in principle makes it easier to integrate an AI scoring tool and follow up statistical properties of both the human and tool using formal logging. This however only applies if the AI tool fits well with the information requirements and workflows of the process.
In the Executive recruitment, the recruiting manager -- who are formally responsible for the vacancy --- were present throughout the entire process. In contrast in the high-volume track the recruiting manager is present only at the beginning and the end, making the chain of responsibility more dependent on the accountability of the process.

It is worth mentioning that from a Fundamental Rights perspective, this process is already high-risk. The Deployer is a large public organization overseeing activities in society-critical functions such as medical care, elderly care, schools and infrastructure. Candidates come from the general public, with wide variation in education, ethnicity, and language background. This means that the affectees of the process are distributed over many vulnerable and protected categories both when it comes to the candidates and their possibility to get occupation and economic stability, but also the beneficiaries of the public activities who are often primarily the vulnerable groups of society. The Deployer being a public organization recognize this responsibility, and have both internal policies and legal frameworks to support accountability --- but it is still worth noting that when doing an in depth analysis using a high-transparency tool like IFM modeling, many of the same questions as arise in AI ethics also arise when looking at purely human decision-procedures.

In the next section we consider the effects of adding automation to steps of these decision processes --- focusing on the high-volume recruitment track.

\subsection{Model Composition and Integration}
\label{sec:integration}

Following the modeling of the Provider’s tool and the Deployer’s recruitment process, IFM was used to construct composite integration models. These capture how decision points, responsibilities, and information availability shift under different configurations—such as replacing interviews, augmenting filtering, or generating ranked shortlists. This enables analysis of how integration choices reshape accountability, oversight, and stakeholder impact.

Like in Section~\ref{sec:deployer}, we focus here on the high-volume case to examine how the same AI system can be integrated at different stages within the same socio-technical structure. %

\begin{figure*}[htbp]
    \centering
    \includegraphics[width=\textwidth]{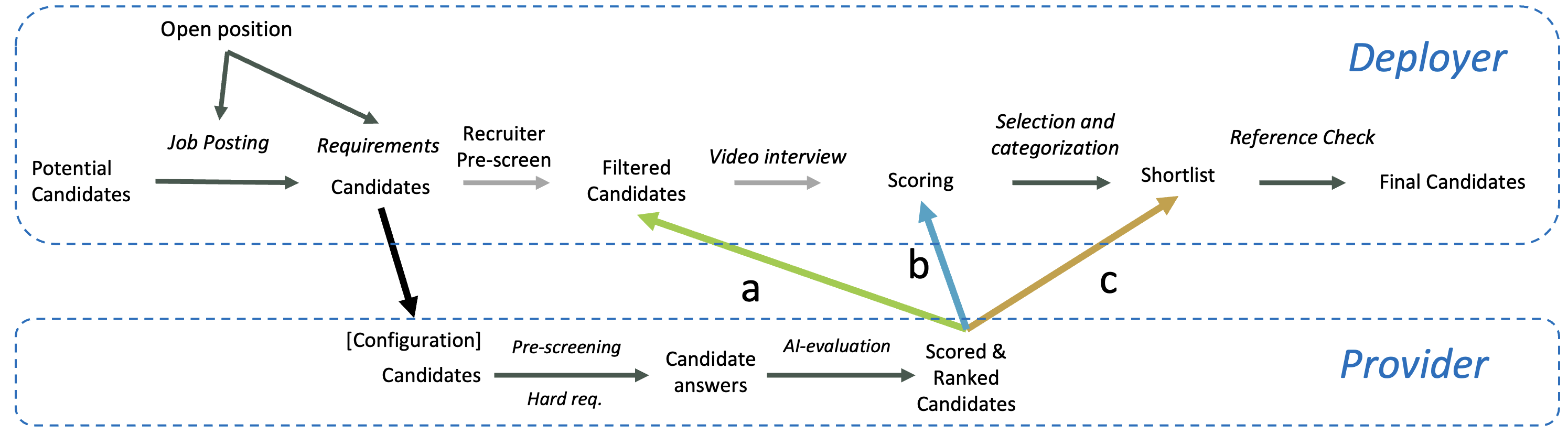}
    
    \caption{Abbreviated information flow model showing three different combination of the Deployer's high-volume recruitment process and the AI tool from Figure \ref{fig:ifm tool} (here displayed in a compressed format). Alternative a) represents a minimal integration where the tool is used for pre-screening only. Alternative b) represents a version where both video-interviews and tool scoring are used in tandem. Alternative c) represents a version where the tool replaces both pre-screening and the video-interview.}
    \label{fig:combined}
\end{figure*}

The high-volume process and the AI system both include stages for hard-requirement screening and candidate scoring. The recruitment process also features a categorization phase—absent in the tool—where candidates are assigned to suitable vacancies based on accumulated information. The categorization plays a significant role in the complete process as candidates are assigned to different potential vacancies depending on the information in the process up to this point.

Based on this structure and suggestions from stakeholders, we identified three integration scenarios with increasing degrees of automation. Each can be examined to determine how different configurations reshape responsibility, user agency, and system oversight:

\begin{enumerate}[label=(\alph*)]
\item \textbf{Minimal integration:} AI used as a soft pre-filter.
\item \textbf{Double evaluation:} AI and recruiter assessments side-by-side.
\item \textbf{Maximal integration:} AI replaces video-interviews and generates shortlists.
\end{enumerate} 
These are summarized in Figure~\ref{fig:combined} and described in detail below.

\paragraph{Minimal integration —-- automation as a soft pre-screening filter.}
In this variant, the tool is configured with a combination of hard rules and scoring questions and used solely to filter candidates before the main recruitment process begins. Candidates who meet all hard requirements pass automatically, while some borderline cases may be retained based on strong scores. After this point, the score is discarded, and the remaining process--—video interviews, evaluation, and sorting—proceeds unchanged. This scenario is described as (a) in Figure~\ref{fig:combined}.

\textit{Analysis:} Because the tool’s outputs do not influence downstream decision-making, the \emph{impact} of any automation bias is structurally limited. However, \emph{automation bias} as a property of the decision process may still be introduced or reinforced—--for instance, if recruiters come to over-rely on early filtering decisions or fail to scrutinize borderline exclusions. In this configuration, traceability is preserved and recruiter agency remains intact, but the added value of the automation may be limited unless the filtering meaningfully improves efficiency or fairness.

\paragraph{Double evaluation —-- automation and human assessment side-by-side.}
In this configuration, the tool is used both for pre-screening and as a scoring component alongside recruiter interviews. Recruiters see both scores, but no procedure defines how to weigh or reconcile them. The human interview may shift to focus on aspects not captured by the tool.
This scenario is described as (b) in Figure~\ref{fig:combined}.

\textit{Analysis:} This setup expands available information but introduces ambiguity in decision logic. Recruiter judgments may still be shaped by automation, allowing \emph{bias} to influence downstream decisions. Because there is no weighting rule or revision mechanism, \emph{impacts} may be harder to detect or contest, and recruiter agency may be reduced if their evaluations become secondary.

\paragraph{Maximal integration —-- automation replaces interviews and shortlists candidates.}
Here, the video interview is removed and the tool performs candidate evaluation and shortlisting. Recruiters make final selections from a list generated solely by automation, without contextual input from interviews.
This scenario is described as (c) in Figure~\ref{fig:combined}.

\textit{Analysis:} This configuration centralizes evaluation within the tool and limits \emph{recruiter agency}. While oversight remains nominally, \emph{bias} can manifest and fully affect outcomes, especially under pressure. Traceability improves technically, but \emph{semantic richness} is lost, weakening fairness and explainability. Responsibility becomes ambiguous when recruiters lack tools or incentives to challenge the output.

Together, these integration scenarios illustrate how IFM can be used to model and compare alternative configurations of automation within the same socio-technical context. By making visible the shifting flow of information, roles, and decision authority, the methodology supports structured analysis of how integration choices reshape responsibility, oversight, and the potential for bias or harm. The following section reflects on the broader implications of these findings and how IFM contributes to the assessment and governance of high-risk AI systems.

\subsection{Project Outcomes}

Three months after the assessment, follow-up interviews with key stakeholders from the municipality indicated that the IFM-based audit had a lasting influence on both organizational understanding and practice. The assessment report was used extensively in internal discussions and planning, supporting management in identifying information gaps and decision dependencies in the existing recruitment workflow. Several concrete changes were implemented, including new formal checkpoints and documentation routines designed to clarify responsibility and information flow throughout the hiring process.

The municipality also incorporated insights from the modeling into internal education initiatives and process development work. Stakeholders described the IFM process as demanding but constructive—its explicit diagrams and discussions exposed organizational dynamics that had previously remained implicit. While the technical language sometimes challenged participants’ sense of familiarity, the participatory modeling sessions were experienced as valuable opportunities for collective reflection.

The method was recognized as not only an analytical tool but also a catalyst for organizational learning. The recruitment side reported continued use of the IFM report for ongoing review and alignment of practices, while the technical side emphasized its value as an educational instrument that revealed how automation and human decision-making interact in practice. The experience has also inspired follow-up activities within the municipality, including initiatives on automation bias and transparency in AI-supported decision systems, as well as interest in further collaborations on the development of audit and governance methods.

\section{Discussion}
\label{sec:discussion}

The case study presented in this paper demonstrates how Information Flow Modeling (IFM) can support analysis of AI system integration in a way that avoids the common pitfalls of monolithic system views and over-abstracted fairness evaluations. Rather than evaluating the AI system or the deployment process in isolation, IFM served as a structural scaffold that made it possible to decompose both into composable, interrelated elements—without losing sight of the broader context in which decisions are made and impacts occur. By modeling information flows, decision points, and stakeholder roles as part of a unified socio-technical structure, IFM makes visible how system behavior and responsibility shift depending on how automation is integrated.

A key strength of the approach is that it supports both granular and holistic reasoning. While the formalism lends itself to structural analysis of decision architectures, it also enables downstream insights into transparency, user responsibility, stakeholder visibility, and ultimately the distribution of impact. This capacity to bridge across the abstraction levels—is crucial in light of critiques that technical tools often flatten the complexity of human systems and displace normative questions into checklist logic or performance metrics~\cite{selbst2019fairness}. In contrast, IFM provides a base layer on which methods like Fundamental Rights Impact Assessments or participatory audits can be built and distributed across system layers. It does not replace these methods but gives them a clearer structural anchoring.

Ultimately, the main contribution of IFM lies not in solving the problems of AI fairness or compliance, but in reshaping what counts as the object of analysis. By moving beyond system-internal evaluations to model information structure and decision formation across human and machine roles, IFM supports assessments that are situated, accountable, and structurally intelligible. In doing so, it helps shift compliance from being an exercise in surface-level validation to one of meaningful structural insight.

\subsection{Reflections on the assessment}

The assessment process repeatedly surfaced a tension between automation bias, the AI system’s \emph{intended use}, and the informational needs of its users. As long as the user is structurally positioned for redundant double-checking, the dynamics of the process will favor automation bias. In contrast, recruiters exercising professional judgment during categorization require fine-grained access to candidate responses and context-specific cues. Reframing the recruiter’s role from passive oversight to active semantic interpretation—grounded in situated decision-making—can support agency and mitigate over-reliance on automation \cite{goddard2012automation}. Information processing tools that support exploration, comparison, and targeted inspection of candidate responses could shift the recruiter’s role from validation to insight generation.

However, this exposes a deeper issue. The stated \emph{intended use} of the Provider in this case --- intentionally or not --- serves both as a promotion of human oversight and a deferral of responsibility. If the tool does not make decisions, yet provides the information on which decisions are based, this cannot fully absolve the Provider from partial responsibility for the outcomes. A more fine-grained understanding of partial or distributed responsibility may be necessary to address such configurations. Alternatively, a more precisely formulated \emph{intended use}, in which the Provider acknowledges responsibility for specific aspects of the decision-making process, could shift the focus from diffuse automation bias to concrete requirements for accuracy, transparency, and accountability.

\subsection{Limitations and Further Work}
As a formal method IFM introduces a technical modeling layer that may present a barrier to adoption. It also relies on interpretation---particularly in settings that use interviews or participatory modeling---as the modeler must translate between informal accounts, technical properties, and socio-organizational context. This requires some familiarity with the domain to be modeled including both the social and technical components. Pedagogic material and software tools are key to address these issues.

Despite these hurdles, the applicability of IFM extends well beyond the specific recruitment setting explored here. As noted in Section~\ref{sec:assessment-methodology}, the framework has already shown potential for use in mapping user agency and could support analysis of transparency or explainability claims in technical systems by tracking access to information across roles and components. A formal treatment of such role-dependent transparency, applied to the same case, is given in~\cite{brannstrom2026answerability}. These qualities make it a promising candidate in both public-sector AI assessments and private-sector development processes—particularly where regulation demands contextual accountability. Further work will explore and further develop these directions.

\section{Conclusion}

This paper has presented a real-world application of the Information Flow Modeling (IFM) methodology as part of an external assessment of a high-risk AI system used in recruitment. The aim was not to evaluate the system itself, but to examine how IFM --- in the role as a structural scaffold --- enables situated, structural reasoning about integration, responsibility, and stakeholder impact. As it allows the construction and composition of both technical and socio-technical models, IFM can form a backbone for understanding the situated effect of automation in both a granular and holistic manner.

In doing so, IFM addresses key challenges in AI accountability and fairness: it counters the abstraction of complex socio-technical processes into simplified technical objects, avoids the framing trap that treats fairness and compliance as internal system properties, and opens integration design to structured exploration rather than checklist validation or crude thought-exercises. By representing human and automated actors within a shared formalism, it provides a bridge between procedural understanding and system-level analysis—supporting transparency, user responsibility and agency and the tracing of stakeholder impacts.

Our application of IFM demonstrates how lightweight modeling tools can reframe the object of analysis --- from isolated technical properties to decision-processes --- thereby revealing real-world structures, risks and effective remedies in AI integration. As AI regulation increases demands on organizations to document and take responsibility for their expanding use of automation, it is crucial that the burden of regulatory compliance is strongly tied to real, visible and transparent positive effects. Without this connection, organizations will gravitate towards tokenistic compliance. We argue that decision-modeling approaches, such as IFM,  strengthen such connections and make assessment results more actionable.

\begin{acks}
This work was supported by the ``AEQUITAS'' project funded by the European Union’s
Horizon Europe research and innovation programme under grant number 101070363.
The views and opinions expressed are those of the author(s) only and do not
necessarily reflect those of the European Union or the granting authority.

The authors thank the auditing firm that conducted the AI system assessment and the
participating municipality for their collaboration and engagement in the case study.
Conflict of interest: author Mattias Brännström is a co-owner of the auditing firm;
the firm was compensated for its consultancy work on the assessment, but not for the
academic research and analysis reported in this paper.
\end{acks}

\bibliographystyle{ACM-Reference-Format}
\bibliography{ref}

@article{barwise1995logic,
  title={On the logic of information flow},
  author={Barwise, Jon and Gabbay, Dov and Hartonas, Chrysafis},
  journal={Logic Journal of IGPL},
  volume={3},
  number={1},
  pages={7--49},
  year={1995},
  publisher={Oxford University Press}
}

@book{barwise1997information,
  title={Information flow: the logic of distributed systems},
  author={Barwise, Jon and Seligman, Jerry and others},
  year={1997},
  publisher={Cambridge University Press}
}

@article{ananny2018seeing,
  title={Seeing without knowing: Limitations of the transparency ideal and its application to algorithmic accountability},
  author={Ananny, Mike and Crawford, Kate},
  journal={new media \& society},
  volume={20},
  number={3},
  pages={973--989},
  year={2018},
  publisher={SAGE Publications Sage UK: London, England}
}

@article{weinberg2022rethinking,
  title={Rethinking fairness: an interdisciplinary survey of critiques of hegemonic ML fairness approaches},
  author={Weinberg, Lindsay},
  journal={Journal of Artificial Intelligence Research},
  volume={74},
  pages={75--109},
  year={2022}
}

@inproceedings{selbst2019fairness,
  title={Fairness and abstraction in sociotechnical systems},
  author={Selbst, Andrew D and Boyd, Danah and Friedler, Sorelle A and Venkatasubramanian, Suresh and Vertesi, Janet},
  booktitle={Proceedings of the conference on fairness, accountability, and transparency},
  pages={59--68},
  year={2019}
}

@inproceedings{metcalf2021algorithmic,
  title={Algorithmic impact assessments and accountability: The co-construction of impacts},
  author={Metcalf, Jacob and Moss, Emanuel and Watkins, Elizabeth Anne and Singh, Ranjit and Elish, Madeleine Clare},
  booktitle={Proceedings of the 2021 ACM conference on fairness, accountability, and transparency},
  pages={735--746},
  year={2021}
}

@misc{EU2024AIAct,
  title        = {Regulation (EU) 2024/1689 of the European Parliament and of the Council of 13 June 2024 laying down harmonised rules on artificial intelligence (Artificial Intelligence Act)},
  howpublished = {\url{https://eur-lex.europa.eu/eli/reg/2024/1689/oj}},
  note         = {Official Journal of the European Union, L 1689, 13 June 2024},
  year         = {2024},
  institution  = {European Union}
}

@article{EC2024Framework,
  author  = {{European Commission}},
  title   = {The EU Artificial Intelligence Act: A Comprehensive Regulatory Framework},
  journal = {Official Journal of the European Union},
  year    = {2024},
  url     = {https://eur-lex.europa.eu/legal-content/EN/TXT/?uri=CELEX%3A32024R1689}
}

@article{VealeBorgesius2021,
  title={Demystifying the Draft EU Artificial Intelligence Act – Analysing the good, the bad, and the unclear elements of the proposed approach},
  author={Veale, Michael and Borgesius, Frederik Zuiderveen},
  journal={Computer Law Review International},
  volume={22},
  number={4},
  pages={97--112},
  year={2021},
  publisher={De Gruyter},
  doi={10.9785/2367-2055-20214},
}

@inproceedings{wieringa2020account,
  title={What to account for when accounting for algorithms: a systematic literature review on algorithmic accountability},
  author={Wieringa, Maranke},
  booktitle={Proceedings of the 2020 conference on fairness, accountability, and transparency},
  pages={1--18},
  year={2020}
}

@inproceedings{green_algorithmic_2020,
  address = {New York, NY, USA},
  title = {Algorithmic Realism: Expanding the Boundaries of Algorithmic Thought},
  url = {https://doi.org/10.1145/3351095.3372840},
  doi = {10.1145/3351095.3372840},
  booktitle = {Proceedings of the 2020 Conference on Fairness, Accountability, and Transparency (FAT* '20)},
  publisher = {Association for Computing Machinery},
  author = {Green, Ben and Viljoen, Salomé},
  year = {2020},
  pages = {19--31}
}

@article{nass2000machines,
  title={Machines and mindlessness: Social responses to computers},
  author={Nass, Clifford and Moon, Youngme},
  journal={Journal of social issues},
  volume={56},
  number={1},
  pages={81--103},
  year={2000},
  publisher={Wiley Online Library}
}

@article{weizenbaum1966eliza,
  title={ELIZA—a computer program for the study of natural language communication between man and machine},
  author={Weizenbaum, Joseph},
  journal={Communications of the ACM},
  volume={9},
  number={1},
  pages={36--45},
  year={1966},
  publisher={ACM}
}

@article{parasuraman1997humans,
  title={Humans and automation: Use, misuse, disuse, abuse},
  author={Parasuraman, Raja and Riley, Victor},
  journal={Human Factors},
  volume={39},
  number={2},
  pages={230--253},
  year={1997},
  publisher={SAGE Publications}
}

@article{dzindolet2003role,
  title={The role of trust in automation reliance},
  author={Dzindolet, Michael T and Peterson, Stephen A and Pomranky, Robert A and Pierce, Lindsey G and Beck, Heather P},
  journal={International Journal of Human-Computer Studies},
  volume={58},
  number={6},
  pages={697--718},
  year={2003},
  publisher={Elsevier}
}

@article{goddard2012automation,
  title={Automation bias: a systematic review of frequency, effect mediators, and mitigators},
  author={Goddard, Kate and Roudsari, Abdul and Wyatt, Jeremy C},
  journal={Journal of the American Medical Informatics Association},
  volume={19},
  number={1},
  pages={121--127},
  year={2012},
  publisher={BMJ Group BMA House, Tavistock Square, London, WC1H 9JR}
}

@inproceedings{raghavan2020mitigating,
  title={Mitigating bias in algorithmic hiring: Evaluating claims and practices},
  author={Raghavan, Manish and Barocas, Solon and Kleinberg, Jon and Levy, Karen},
  booktitle={Proceedings of the 2020 conference on fairness, accountability, and transparency},
  pages={469--481},
  year={2020}
}

@inproceedings{binns2018fairness,
  title={Fairness in machine learning: Lessons from political philosophy},
  author={Binns, Reuben},
  booktitle={Conference on fairness, accountability and transparency},
  pages={149--159},
  year={2018},
  organization={PMLR}
}

@inproceedings{raji2020closing,
  title={Closing the AI Accountability Gap: Defining an End-to-End Framework for Internal Algorithmic Auditing},
  author={Raji, Inioluwa Deborah and Smart, Daisy and White, Rebecca and Mitchell, Margaret and Gebru, Timnit and Hutchinson, Ben and Smith-Loud, Jamila and Theron, Daniel and Barnes, Parker},
  booktitle={Proceedings of the 2020 ACM Conference on Fairness, Accountability, and Transparency},
  pages={33--44},
  year={2020},
  publisher={ACM}
}

@article{cobbe2021reviewable,
  title={Reviewable Automated Decision-Making: A Framework for Accountable Algorithmic Systems},
  author={Cobbe, Jennifer and Singh, Jatinder and Whittaker, Zachary},
  journal={Computer Law \& Security Review},
  volume={41},
  pages={105567},
  year={2021},
  publisher={Elsevier}
}

@article{lopez2021bias,
  title={Bias Amplification in Artificial Intelligence Systems},
  author={Lopez, Nathalie A},
  journal={AI \& Society},
  volume={36},
  pages={551--559},
  year={2021},
  publisher={Springer}
}

@article{cobbe2023supply,
  title={The Algorithmic Supply Chain},
  author={Cobbe, Jennifer and Singh, Jatinder},
  journal={Philosophy \& Technology},
  volume={36},
  number={1},
  pages={16},
  year={2023},
  publisher={Springer}
}

@inproceedings{dignum2005omni,
  title={Organizing Multiagent Systems},
  author={Dignum, Virginia and Dignum, Frank},
  booktitle={Proceedings of the Fourth International Joint Conference on Autonomous Agents and Multiagent Systems (AAMAS 2005)},
  pages={organized-systems--organization-model},
  year={2005},
  organization={ACM}
}

@article{novelli2023taking,
  title={Taking AI Risks Seriously: A Scenario-Based Approach to AI Risk Management and Assessment},
  author={Novelli, Claudio and Rizza, Davide and van Kleek, Max and Shadbolt, Nigel},
  journal={AI and Ethics},
  year={2023},
  publisher={Springer},
  doi={10.1007/s43681-023-00298-3}
}

@misc{novelli2024scenario,
  title={AI Risk Assessment: A Scenario-Based Proportional Methodology for the AI Act},
  author={Novelli, Claudio and van Kleek, Max and Shadbolt, Nigel},
  year={2024},
  howpublished={arXiv preprint arXiv:2402.03653},
  note={Available at \url{https://arxiv.org/abs/2402.03653}}
}

@misc{ifm2025guide,
  author       = {Br{\"a}nnstr{\"o}m, Mattias and Xanthopoulou, Themis Dimitra},
  title        = {Information Flow Model ({IFM}) -- Methodological Guide},
  year         = 2025,
  doi          = {10.5281/zenodo.16252143},
  publisher    = {Zenodo}
}

@inproceedings{brannstrom2023impact,
  author    = {Br{\"a}nnstr{\"o}m, Mattias and Jiang, Lili and Aler Tubella, Andrea and Dignum, Virginia},
  title     = {Impact Based Fairness Framework for Socio-Technical Decision Making},
  booktitle = {Proceedings of the 1st Workshop on Fairness and Bias in AI (AEQUITAS 2023) co-located with ECAI 2023},
  series    = {CEUR Workshop Proceedings},
  volume    = {3523},
  year      = {2023},
  url       = {https://ceur-ws.org/Vol-3523/paper12.pdf}
}

@misc{brannstrom2026answerability,
  author = {Br{\"a}nnstr{\"o}m, Mattias and Xanthopoulou, Themis Dimitra and Jiang, Lili},
  title  = {Transparency as Answerability: Role-Dependent Information Flow Modeling},
  year   = {2026},
  note   = {Under revision at ACM Journal on Responsible Computing}
}

\end{document}